\documentclass[10pt,conference]{IEEEtran}
\IEEEoverridecommandlockouts
\usepackage{cite}
\usepackage{amsmath,amssymb,amsfonts}
\usepackage{algorithmic}
\usepackage{graphicx}
\usepackage{textcomp}
\usepackage{xcolor}
\usepackage{booktabs}
\usepackage{caption}
\usepackage{multirow}
\def\BibTeX{{\rm B\kern-.05em{\sc i\kern-.025em b}\kern-.08em
    T\kern-.1667em\lower.7ex\hbox{E}\kern-.125emX}}

\begin{document}

\title{Unified Long Video Inpainting and Outpainting via Overlapping High-Order Co-Denoising}
\author{\IEEEauthorblockN{Shuangquan Lyu}
\IEEEauthorblockA{\textit{Carnegie Mellon University}}
\and
\IEEEauthorblockN{Jian Mao}
\IEEEauthorblockA{\textit{Jilin University}}
\and
\IEEEauthorblockN{Yue Ma}
\IEEEauthorblockA{\textit{Tsinghua University}}
}

\twocolumn[{%
\renewcommand\twocolumn[1][]{#1}%
\maketitle
\vspace{-0.9cm}
\begin{center}
    \centering
    \captionsetup{type=figure}
    \includegraphics[width=0.9\linewidth]{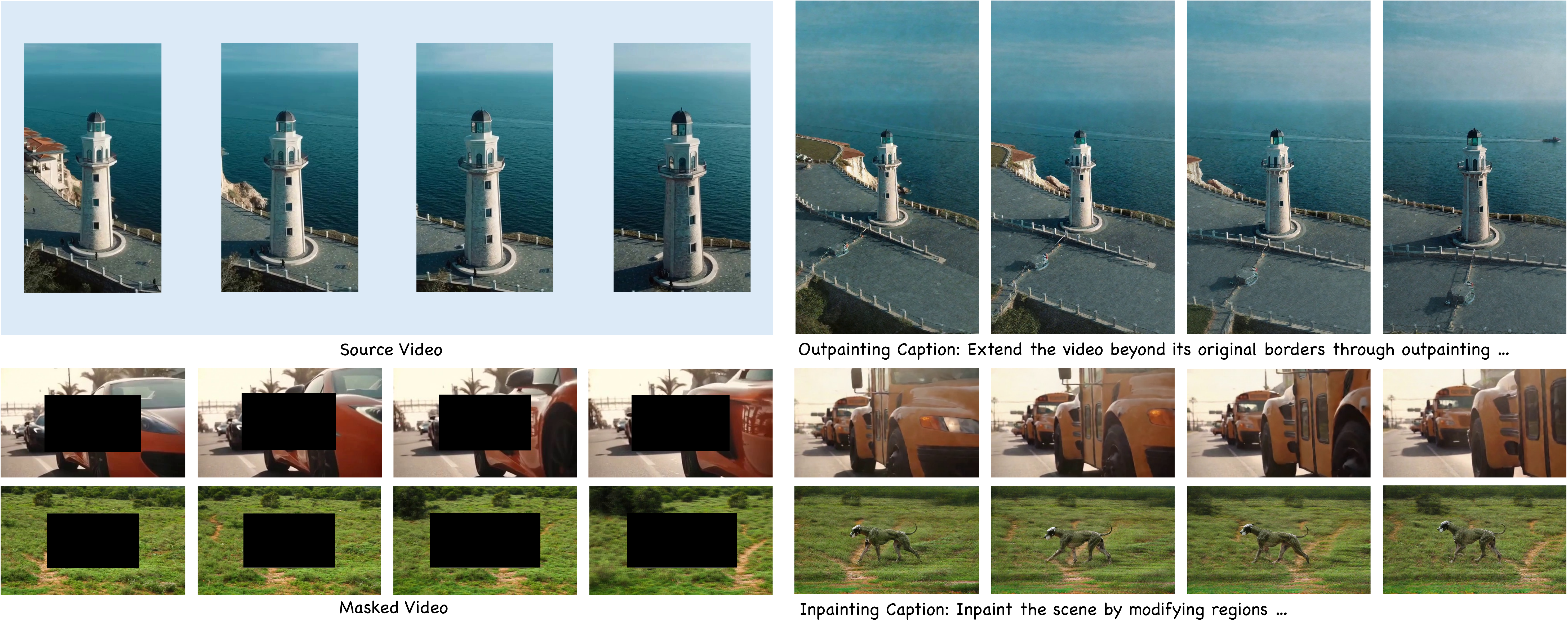}
    \vspace{-6pt}
    \captionof{figure}{Showcase of our methods. We present a unified approach for long video inpainting and outpainting that extends a text-to-video diffusion model to generate long-horizon, spatially edited videos with high fidelity under memory bounded by window size.`}
    \label{fig:teaser}
    \vspace{3pt}
\end{center}%
}]
    
\begin{abstract}
Diffusion-based text-to-video models are increasingly capable, but mask-based editing over hundreds of frames remains challenging: naïve long-video generation suffers from memory blow-up, window seams, and temporal drift, while existing editors often require specialized modules or heavy fine-tuning. We present Overlapping High-Order Co-Denoising, a lightweight framework that turns a single pre-trained text-to-video model into a unified inpainting--outpainting editor. We train only LoRA adapters using mixed interior and border masks together with a dual-region loss that improves synthesis inside the mask while explicitly preserving known content. At inference, we denoise long latent sequences using overlapping windows, apply second-order Heun sampling within each window, and fuse overlaps with Hamming-weighted blending to reduce boundary artifacts and improve temporal coherence. On InpaintBench (30 real-world videos, 81--300 frames), our method outperforms Wan 2.1 variants and VACE in background faithfulness (SSIM/LPIPS), temporal consistency (tLPIPS), and text alignment (CLIP), and scales to long horizons, demonstrated up to 800 frames, with memory bounded by the chosen window size.
\end{abstract}


\section{Introduction}
Generating videos from textual descriptions is a fundamental problem in generative modeling. Recent foundation text-to-video diffusion models~\cite{kong2024hunyuanvideo,wan2025wan} have made remarkable progress in generating short video clips from textual descriptions. With the large amount of training data, large-scale training and architectural refinements, they naturally possess video editing capabilities. Inpainting and outpainting~\cite{Perazzi_CVPR_2016, xu2018youtube} are two video editing tasks which have been widely studied. Building on these foundation models, recent work has explored text-guided masked editing.

However, two major challenges remain when adapting a single model to long-video editing. First, extending diffusion models to substantially longer videos is difficult due to memory limits and temporal consistency issues---naïve scaling often leads to drift or stitching artifacts over time. Second, enabling fine-grained mask-based inpainting and outpainting within one unified framework is non-trivial. While recent works have begun exploring controllable video editing, current foundation models still offer limited spatial control without additional specialized modules. Enhancing a pre-trained model’s controllability for selective region editing, especially over long durations, remains an open problem.

Built on Wan~\cite{wan2025wan}, we adapt a single pre-trained text-to-video model for \emph{mask-based} inpainting and outpainting by inserting LoRA adapters and training on randomly masked clips with a dual-region MSE loss. For long-horizon inference, we denoise overlapping temporal windows using a second-order Heun update and fuse overlaps with Hamming-weighted blending to reduce seams and drift while keeping memory bounded by window size. Our design improves long-video fidelity and temporal stability while keeping GPU memory bounded. Experiments show consistent gains on CLIP~\cite{radford2021learning}, SSIM~\cite{wang2004SSIM}, LPIPS~\cite{zhang2018lpips} and tLPIPS over strong baselines (e.g., Wan 2.1 14B).

Our contributions are summarized as follows:
\begin{itemize}
    \item We adapt a single pre-trained text-to-video diffusion model to both video inpainting and outpainting using mask-conditioned LoRA fine-tuning with a dual-region reconstruction loss.
    \item We develop a sliding-window diffusion inference scheme that combines a Heun solver with Hamming-weighted blending, reducing seam artifacts and temporal drift while keeping memory bounded by window size.
    \item We build and evaluate on InpaintBench (30 videos, 81--300 frames), achieving consistent improvements in CLIP, SSIM, LPIPS, and tLPIPS over strong baselines while enabling long-horizon editing, demonstrated up to 800 frames, with bounded memory.
\end{itemize}
We unify mask-conditioned LoRA adaptation and long-horizon co-denoising in a single pipeline. The dataset and code will be released publicly.

\begin{figure*}[t!]
    \centering
    \includegraphics[width=0.9\textwidth]{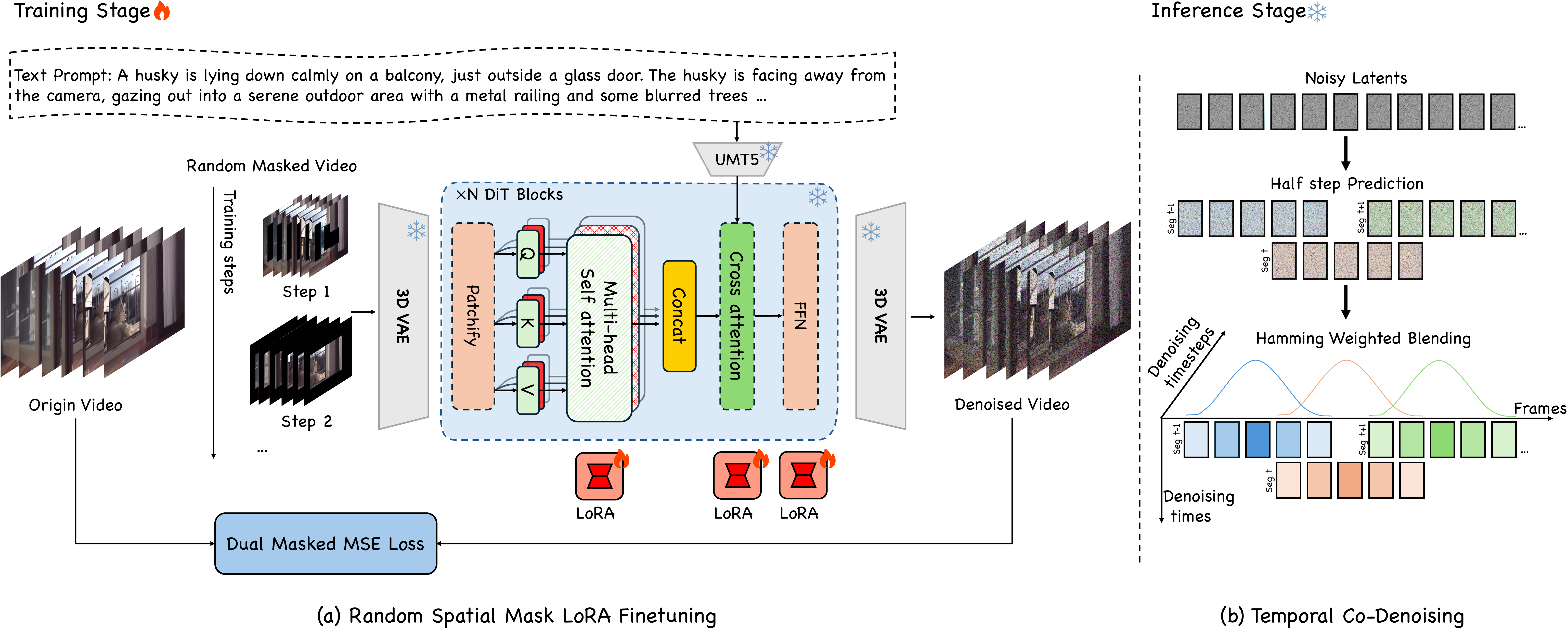}
    \vspace{-1em}
    \caption{\textbf{Overview.} We introduce a unified LoRA-based fine-tuning pipeline for both video inpainting and outpainting on InpaintBench. During training, each clip is randomly masked with either (i) \emph{border masks}, which remove frame boundaries, or (ii) \emph{interior masks}, which occlude regions inside the frame; a dual-region MSE loss then encourages accurate hole filling while preserving unmasked content. At inference, we partition long sequences into overlapping windows and perform temporal co-denoising using a two-stage Heun sampler with Hamming-weighted blending, yielding smoother long-video editing with reduced boundary artifacts.}
    \label{fig:pipeline}
    \vspace{-1em}
\end{figure*}

\section{Related Work}
\noindent\textbf{Foundation text-to-video diffusion models.}
Recent large-scale text-to-video diffusion models produce high-quality short clips with strong motion priors and text alignment \cite{kong2024hunyuanvideo,wan2025wan}. However, they are typically trained on fixed temporal horizons and offer limited \emph{native} support for long-horizon, mask-based region edits, making naive extensions prone to drift and boundary artifacts.

\noindent\textbf{Long video generation and extension.}
Training diffusion models directly on long videos is computationally expensive, so most text-to-video models are limited to short clips and suffer quality degradation when naively extended. Autoregressive approaches generate variable-length videos but often accumulate errors over time~\cite{he2022latent, ma2024followpose, feng2025dit4edit, ma2025followcreation, shen2025follow, long2025follow, ma2026fastvmt, ma2025followyourmotion, ma2025controllable, ma2026group, henschel2025streamingt2v, villegas2022phenaki}. A large body of work instead explores \emph{tuning-free} strategies to extend off-the-shelf diffusion models at inference time, including temporal co-denoising, noise rescheduling, and attention or latent blending~\cite{kim2024fifo, qiu2023freenoise, wang2023gen, cai2025ditctrl}. Notably, Gen-L-Video~\cite{wang2023gen} introduces a windowed temporal co-denoising framework, while FreeNoise~\cite{qiu2023freenoise} and DiTCtrl~\cite{cai2025ditctrl} improve long-range consistency via noise and attention manipulation. In this paper, we build on the windowed co-denoising paradigm and incorporate high-order integration together with Hamming-weighted blending to further reduce seam artifacts and temporal drift.

\noindent\textbf{Diffusion-based video inpainting and outpainting.}
Classical video inpainting benchmarks (e.g., DAVIS, YouTube-VOS) assume pixel-level ground truth for missing regions and typically emphasize temporal propagation. More recent diffusion-based approaches incorporate text guidance and masked editing. VACE~\cite{jiang2025vace} provides a unified video creation/editing framework that supports masked edits, making it a strong baseline for our V2V setting. AVID~\cite{zhang2024avid} introduces a MultiDiffusion-style strategy to support any-length video inpainting, and Follow-Your-Canvas~\cite{chen2024followyourcanvas} and Be-Your-Outpainter~\cite{wang2024beyouroutpainter} focus on large-extent/high-resolution outpainting via windowed fusion and input-specific adaptation. VideoPainter~\cite{bian2025videopainter} enables plug-and-play mask-based video editing with pre-trained diffusion models. In contrast to methods that introduce specialized modules or focus on a single edit mode, we emphasize a lightweight \emph{unified} pipeline that adapts a single T2V backbone to both interior-hole inpainting and border outpainting, while scaling to long videos through a sampler designed to suppress window seams and temporal drift.

\noindent\textbf{Parameter-efficient adaptation with LoRA.}
Adapting billion-parameter diffusion backbones via full fine-tuning is costly. LoRA~\cite{hu2022lora} enables parameter-efficient specialization by learning low-rank updates to frozen weights, and has become a standard tool for quickly adapting foundation models. We leverage LoRA to inject \emph{mask-controllable} behavior into a pre-trained T2V model with minimal trainable parameters, enabling both inpainting and outpainting under a shared training objective. 

\noindent\textbf{High-order diffusion sampling.}
Diffusion sampling can be interpreted as numerically integrating a reverse-time process, where solver choice affects stability and error accumulation~\cite{ho2020denoising}. Higher-order updates such as Heun-style steps can improve sample quality per function evaluation and reduce discretization error~\cite{karras2022elucidating}. Building on these insights, we integrate a two-stage Heun update into windowed temporal co-denoising and combine it with Hamming-weighted overlap blending to mitigate seam artifacts and long-horizon drift in edited videos.


\section{Methodology}

Our approach comprises three components: (1) LoRA-based spatial mask fine-tuning, (2) inference-time mask conditioning, and (3) long-horizon temporal co-denoising with a high-order solver. Together, these elements enable a single foundation model (Wan 2.1 14B) to perform both inpainting and outpainting within a unified framework. Figure \ref{fig:pipeline} illustrates the overall pipeline.

\subsection{LoRA-based Spatial Mask Fine-Tuning}\label{sec:lora}
Although the original Wan model can perform basic video editing, its limited controllability makes more complex mask-guided edits difficult. To enable mask-conditioned generation without modifying the core DiT architecture, we inject LoRA adapters into self-attention, cross-attention, and feed-forward blocks. Concretely, for every weight matrix $W \in \mathbb{R}^{d\times k}$ in the frozen DiT, we learn a residual update:
\begin{equation}
  \Delta W = B A, \quad B\in\mathbb{R}^{d\times r},\; A\in\mathbb{R}^{r\times k},
\end{equation}
where $r\ll \min(d,k)$ is the LoRA rank. The adapted weight is $W^* = W + \Delta W$, and thus we optimize only $A$ and $B$ during training time instead of the entire DiT blocks.

During each iteration, we sample a video clip $\{x_f\}_{f=1}^T$ and randomly apply one of two mask types to all frames: (1) \emph{Border masks}, where pixels outside a central rectangle covering $\alpha \in [0.5,0.8]$ of each spatial dimension are masked out to simulate outpainting; and (2) \emph{Interior masks}, where $m \sim \mathrm{Uniform}\{1,4\}$ rectangular regions within each frame are masked to simulate inpainting. Sampling both mask types during training enables a single model to learn both inpainting and outpainting within one unified framework.


We distinguish the video frame index $f \in \{1,\dots,T\}$ from diffusion timesteps. Let $x_f \in \mathbb{R}^{H \times W \times 3}$ denote the ground-truth frame and $\hat{x}_f$ denote the reconstructed output frame after VAE decoding. Let $M_f \in \{0,1\}^{H \times W}$ be the binary mask for frame $f$, where $M_f(u,v)=1$ indicates pixels to be synthesized and $M_f(u,v)=0$ indicates known pixels.

We define masked-region and unmasked-region reconstruction losses as
\begin{align}
  L_{\mathrm{masked}} &=
  \sum_{f=1}^{T} \left\| M_f \odot (\hat{x}_f - x_f) \right\|_2^2, \\
  L_{\mathrm{unmasked}} &=
  \sum_{f=1}^{T} \left\| (1 - M_f) \odot (\hat{x}_f - x_f) \right\|_2^2 .
\end{align}
The final dual-region loss is a weighted sum:
\begin{equation}
  L_{\mathrm{Dual}} =
  \lambda \, L_{\mathrm{masked}} + (1-\lambda)\, L_{\mathrm{unmasked}},
\end{equation}
where $\lambda \in [0,1]$ balances hole-filling fidelity and context preservation. We empirically set $\lambda=0.9$ in all experiments unless noted otherwise.

\subsection{Inference-Time Mask Conditioning}
At inference, we use the same masking operator as in training: known pixels are provided through masked latents, while the model synthesizes masked regions guided by the text prompt and LoRA adaptation. This preserves known content outside the edit region while allowing synthesis inside the mask.

\paragraph{Inpainting}
Given an input video $\{x_f\}_{f=1}^{T}$ and a user-specified mask $\{M_f\}_{f=1}^{T}$, we form masked frames
\begin{equation}
  \tilde{x}_f = (1 - M_f) \odot x_f, \qquad f=1,\dots,T,
\end{equation}
and encode them with the Wan VAE encoder $E(\cdot)$:
\begin{equation}
  z_f' = E(\tilde{x}_f).
\end{equation}
The LoRA-adapted diffusion model then performs conditional denoising using $\{z_f'\}$ and the text prompt.

We do \emph{not} explicitly clamp (re-impose) known pixels at each denoising step. Instead, preservation of unmasked content is encouraged by the dual-region loss (Sec.~\ref{sec:lora}), which penalizes deviations in $(1-M_f)$ regions during fine-tuning. This avoids hard transitions that can arise from repeatedly overwriting known pixels during denoising.

\paragraph{Outpainting}
To outpaint beyond the original field of view, we expand the latent canvas by padding. Let $P(\cdot)$ denote zero-padding to a user-specified spatial size. For each frame latent $z_f$, we form
\begin{equation}
  z_f' = P(z_f),
\end{equation}
and apply the same conditional diffusion process to synthesize content in the padded area. The padded latent canvas exposes new spatial regions for synthesis while retaining the original video content as context for coherent scene extension.

\subsection{Arbitrary-Length Temporal Co-Denoising}
Since Wan is trained on fixed short sequences, naively processing a longer video of $T \gg W$ frames incurs severe memory growth and temporal consistency issues. We therefore adopt a sliding window co-denoising strategy with adjustable overlap length $O$ and Hamming-weighted blending at each diffusion time step $t$. The process is illustrated in Figure \ref{fig:pipeline} (b). The key idea is to process a long video as overlapping temporal segments, denoise each segment locally, and then fuse the overlapping predictions smoothly.

\begin{figure*}[t]
    \vspace{-0.9cm}
    \centering
    \includegraphics[width=0.9\textwidth]{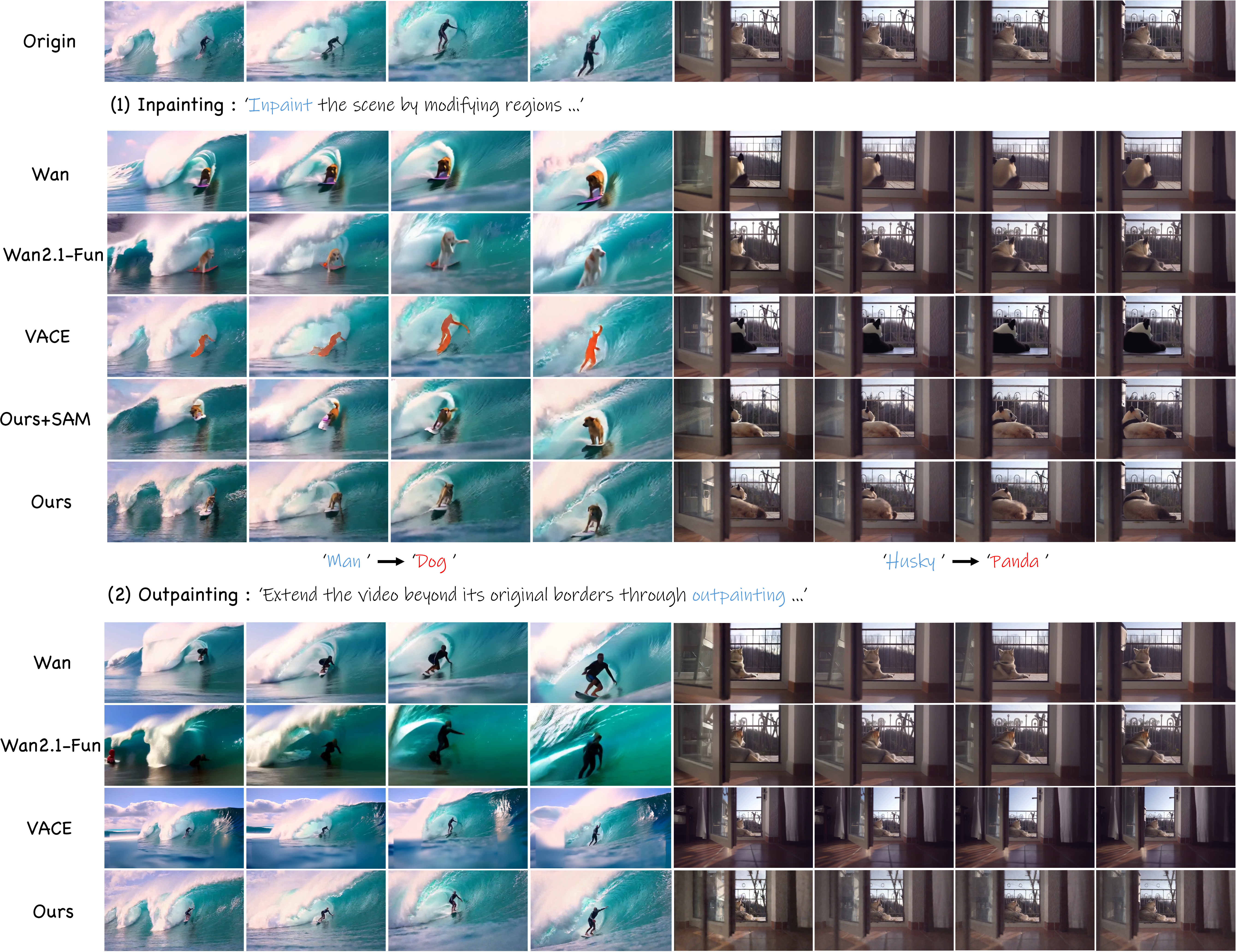}
    \caption{\textbf{Qualitative results.} Inpainting (top) and outpainting (bottom) on long videos. Our method better preserves structure and motion while reducing boundary seams and temporal drift relative to Wan 2.1 variants and VACE.}
    \label{fig:qualitative}
    \vspace{-1em}
\end{figure*}

\paragraph{Window extraction}
Let $Z^{(n)} \in \mathbb{R}^{T \times d}$ denote the full latent sequence at diffusion step $n$ (with $n=N$ the initial noisy latent and $n=0$ the final latent). We extract overlapping windows using start indices
\begin{equation}
  s_i = 1 + (i-1)(W-O), \qquad i = 1,\dots,I,
\end{equation}
where
\begin{equation}
  I = \max\left(1, \left\lceil \frac{T-W}{W-O} \right\rceil + 1 \right).
\end{equation}
The $i$-th window at step $n$ is
\begin{equation}
  Z^{(n)}_i = Z^{(n)}[\,s_i : s_i + W - 1\,] \in \mathbb{R}^{W \times d}.
\end{equation}

\paragraph{Two-stage Heun update (within each window)}
Let $\{\sigma_n\}_{n=0}^{N}$ be the discrete noise schedule and $\Delta\sigma_n = \sigma_{n-1} - \sigma_{n}$. We denote by $\Phi(\cdot,\sigma)$ the diffusion-model-derived update direction (e.g., the ODE derivative in EDM-style sampling). In our implementation, $\Phi(\cdot,\sigma)$ is computed from the model’s noise-prediction output under the standard EDM ODE formulation. For each window $i$, we perform a second-order Heun step:
\begin{align}
  K_1 &= \Phi\!\left(Z^{(n)}_i, \sigma_n\right), \\
  \tilde{Z}_i &= Z^{(n)}_i + \Delta\sigma_n \, K_1, \\
  K_2 &= \Phi\!\left(\tilde{Z}_i, \sigma_{n-1}\right), \\
  \hat{Z}^{(n-1)}_i &= Z^{(n)}_i + \Delta\sigma_n \, \frac{K_1 + K_2}{2}.
\end{align}
This requires two model evaluations per step and reduces error accumulation over long sequences. Compared with a first-order update, Heun sampling reduces local discretization error and improves temporal stability.

\paragraph{Overlap blending with Hamming weights}
We blend overlapping windows using a 1D Hamming window of length $W$:
\begin{equation}
  w_j = \alpha - \beta \cos\!\left(\frac{2\pi (j-1)}{W-1}\right),
  \qquad j=1,\dots,W,
\end{equation}
with $\alpha=0.54$ and $\beta=0.46$.
For each frame index $f \in \{1,\dots,T\}$, we aggregate all windows covering $f$:
\begin{equation}
  Z^{(n-1)}[f] =
  \frac{
    \sum_{i:\, f \in [s_i,\, s_i+W-1]}
    w_{f-s_i+1}\, \hat{Z}^{(n-1)}_i[\,f-s_i+1\,]
  }{
    \sum_{i:\, f \in [s_i,\, s_i+W-1]} w_{f-s_i+1}
  }.
\end{equation}
Hamming weights down-weight window boundaries and emphasize segment centers, where predictions are typically more stable, reducing seams relative to naïve stitching or uniform averaging.

\paragraph{Complexity and Parallelism} 
At each diffusion step, we process windows sequentially (or in parallel if multiple devices are available) and accumulate their contributions into a global buffer. Memory is bounded by the window size $W$, and runtime scales linearly with video length $T$ up to the constant factor from overlapping windows and the two-stage solver.

After all steps to $t=0$, we decode $X_0$ using the VAE decoder to obtain the final video frames. Our ablation study (Section~\ref{sec:ablations}) confirms the importance of LoRA fine-tuning and Heun sampling, while qualitative comparisons suggest that Hamming blending further helps suppress boundary artifacts in long-video editing.

\section{Experiments}

\subsection{Datasets}
Existing benchmarks such as DAVIS~\cite{Perazzi_CVPR_2016} and YouTube-VOS~\cite{xu2018youtube} focus on short clips. To evaluate long-form editing, we construct \textbf{InpaintBench}, a collection of 30 publicly available real-world videos spanning human actions, animals, outdoor scenes, and indoor environments, with lengths from 81 to 300 frames.
For each video, we provide a base caption, an \emph{inpainting} prompt, an \emph{outpainting} prompt, and a mask specification used consistently across methods. We release the videos, prompts, and mask generators for reproducible evaluation. Despite its modest size, the benchmark covers diverse content and long-horizon editing scenarios underrepresented in short-clip benchmarks.

\subsection{Implementation Details}
Our method is built on the official DiffSynth-Studio~\cite{diffsynthstudio} codebase and the publicly available 14B-parameter Wan 2.1 text-to-video model. We inject LoRA adapters (rank 16) into all self-attention layers, cross-attention layers, and feed-forward sublayers. Training is performed on a single NVIDIA H100 GPU using the AdamW optimizer with a fixed learning rate of $1\times10^{-4}$. We train for 2000 steps on a curated set of 25 video clips spanning human activities, animals, indoor scenes, and outdoor scenes. Each sample is temporally resampled to 81 frames at $416 \times 240$. Border and interior masks are generated on the fly to simulate outpainting and inpainting. Training takes about one hour. Text prompts are generated with ChatGPT, then manually reviewed and refined before pairing with each clip. At inference, we adopt a two-stage Heun solver and Hamming-weighted blending, together with classifier-free guidance, for text-guided video-to-video generation.

\subsection{Evaluation Setup}

\textbf{Tasks.} We evaluate two editing tasks:
(1) Object inpainting: adding or replacing objects within videos, and
(2) Scene outpainting: extending content beyond original frame boundaries.

\textbf{Baselines.} We compare against Wan 2.1 14B~\cite{wan2025wan}, Wan2.1-Fun-Control~\cite{alibaba_pai_wan21_fun14b_control}, and VACE~\cite{jiang2025vace}, which represent strong recent text-to-video and video-to-video diffusion systems supporting conditional generation. While systems like AVID~\cite{zhang2024avid} and VideoPainter~\cite{bian2025videopainter} address video completion, they rely on architectures or inference mechanisms that differ substantially from our lightweight LoRA-adaptation setting. In contrast, our method creates a lightweight, trainable adapter that is structurally simpler and more scalable. We compare against VACE and Wan-Control as they represent the state-of-the-art in tuning-based and control-based foundation models, which are the direct counterparts to our proposed LoRA approach.
Unless explicitly modified, each baseline uses its default inference sampler and schedule from the corresponding public implementation. The effect of our two-stage Heun solver is isolated separately in the ablation study.

\begin{figure*}[t!]
    \centering
    \includegraphics[width=0.85\textwidth]{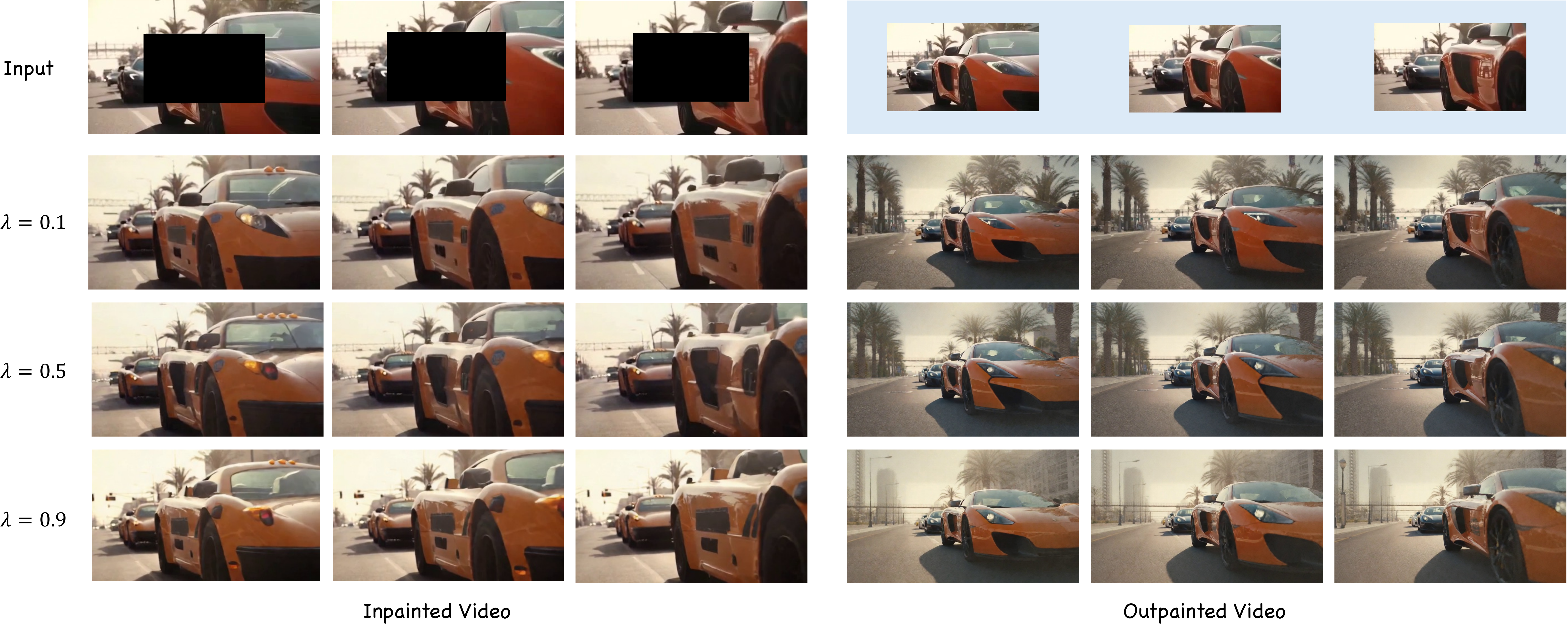}
    \caption{\textbf{Effect of $\lambda$ in $L_\mathrm{Dual}$.} Smaller $\lambda$ preserves context but under-fills; larger $\lambda$ improves completion but may perturb known regions. $\lambda{=}0.9$ gives the best trade-off.}
    \label{fig:ablation_weight}
    \vspace{-1em}
\end{figure*}

\textbf{Evaluation Protocol.} 
In our object replacement and scene extension tasks, edited regions have no pixel-level ground truth, so we separate \emph{background preservation} from \emph{edit realism/alignment}. We therefore evaluate edited content using prompt alignment and temporal consistency rather than pixel reconstruction.
(1) \textbf{Background preservation / leakage:} we compute SSIM (Structural Similarity)~\cite{wang2004SSIM} and LPIPS (Learned Perceptual Image Patch Similarity)~\cite{zhang2018lpips} \emph{only on the unmasked region} $(1-M)$ between the input and edited videos, which directly measures whether methods inadvertently alter known pixels outside the target area.
(2) \textbf{Prompt alignment in edited regions:} we report CLIP (Contrastive Language--Image Pre-training) similarity~\cite{radford2021learning} between the text prompt and generated frames, computed on the \emph{masked region crop} (bounding box of $M$) to focus on the intended edit rather than the preserved background.\footnote{If the masked region is disconnected (multi-rectangle), we take the tight union bounding box; details in code.}
(3) \textbf{Temporal stability:} we report tLPIPS as the average LPIPS between consecutive generated frames, which captures flicker and short-term inconsistency (lower is better).

We report metrics averaged over videos and frames; for Table~\ref{tab:overall_quantitative} we average each metric across both tasks with equal weight unless stated otherwise. This protocol avoids invalid pixel metrics inside the edited region while still quantifying (i) faithfulness to preserved content, (ii) semantic alignment of the synthesized content, and (iii) temporal smoothness.


\begin{table}[t] 
  \centering
  \begin{tabular}{@{} l  c c c c @{}}
    \toprule
    Method      & {CLIP$\uparrow$} & {SSIM$\uparrow$} & {LPIPS$\downarrow$} & {tLPIPS$\downarrow$} \\
    \midrule
    Wan2.1 14B  & 29.2 & 0.563 & 0.130 & 0.225\\
    Wan2.1-Fun-Control 14B  & 27.1 & 0.589 & 0.162  & 0.228\\
    VACE        & 28.4 & 0.612 & 0.125 & 0.227\\
    Ours        & \textbf{29.4}  & \textbf{0.613}   & \textbf{0.121}  & \textbf{0.224}\\
    \bottomrule
  \end{tabular}
\caption{Quantitative comparisons on InpaintBench. Results are reported as [mean $\pm$ standard deviation across videos / mean with 95\% bootstrap confidence intervals]. $\uparrow$ indicates higher is better and $\downarrow$ indicates lower is better. SSIM and LPIPS are computed on the unmasked region $(1-M)$ for both inpainting and outpainting, measuring background preservation. CLIP evaluates prompt alignment of edited regions, and tLPIPS measures temporal consistency.}
\label{tab:overall_quantitative}
\end{table}

\textbf{Qualitative results.} Figure~\ref{fig:qualitative} shows representative inpainting and outpainting examples (77--181 frames). Compared with Wan 2.1 variants and VACE, our method better maintains motion continuity, reduces boundary seams, and produces more coherent prompt-aligned edits. See the supplementary videos for full temporal comparisons.

\subsection{Ablation Studies}\label{sec:ablations}
We conduct ablations to measure the contribution of each component and the selection of hyperparameters.

\noindent
\paragraph{\textbf{Selection of \(\lambda\) in \(L_\mathrm{Dual}\)}}  
We evaluated \(\lambda\in\{0.1,0.5,0.9,1.0\}\) and report the results in Table~\ref{tab:loss-weight-quantitative}. Increasing \(\lambda\) places greater emphasis on masked-region reconstruction, which improves hole filling but introduces slightly larger deviations in the unmasked areas. The best balance of CLIP and SSIM is achieved at \(\lambda=0.9\), as confirmed by both quantitative metrics and the qualitative examples in Figure~\ref{fig:ablation_weight}. In practice, \(\lambda\) can be tuned along with other hyperparameters to match specific application requirements.

\begin{table}[t]
  \centering
  \footnotesize
  \setlength{\tabcolsep}{3.5pt}
  \begin{tabular}{@{}ccc|ccc@{}}\toprule
 & \multicolumn{2}{c}{Inpainting} &   \multicolumn{2}{c}{Outpainting}\\
    \midrule
    $\lambda$ & {CLIP$\uparrow$} &  {SSIM$\uparrow$} & {CLIP$\uparrow$}  &{SSIM$\uparrow$} \\
    0.1 & 23.4 & 0.695 & 22.9 & 0.672 \\
    0.5 & 26.7 & 0.661 & 25.8 & 0.645 \\
    0.9 & 29.8 & 0.636 & 28.1 & 0.592 \\
    1.0 & 29.6 & 0.628 & 28.0 & 0.587 \\
    \bottomrule
  \end{tabular}
\caption{$L_{\mathrm{masked}}$ weight $\lambda$ impact on performance}
\label{tab:loss-weight-quantitative}
\end{table}
\noindent
\textbf{Arbitrary-length Temporal Co-denoising}. To assess the limitations of other approaches that encode and generate entire videos simultaneously, we conducted experiments to test the upper limit of video lengths on 1 NVIDIA H100 80GB GPU. We encounter out-of-memory errors when the generation video of the same size is longer than the frames' upper limit.
\begin{table}[htbp]
  \centering
  \begin{tabular}{@{} l  c @{}}
    \toprule
    Method     & Max number of frames  \\
    \midrule
    VACE   & fixed 81   \\
    Wan 2.1 14B   & max 245  \\
    \textbf{Ours} & bounded by window size; demonstrated up to 800 frames   \\
    \bottomrule
  \end{tabular}
  \caption{Supported maximum video length at 1600×800 resolution.}
  \label{tab:length-comparison}
\end{table}

\noindent\textbf{Two‐stage Heun solver}. Table~\ref{tab:heun-solver} compares video generation with and without our two-stage Heun solver. Incorporating Heun’s method raises SSIM, reduces LPIPS and tLPIPS. These gains demonstrate that the high-order solver substantially improves both reconstruction fidelity, perceptual quality and temporal consistency.

\begin{table}[htbp]
  \centering
  \begin{tabular}{@{} l  c c c c@{}}
    \toprule
    Method       & {SSIM$\uparrow$} & {LPIPS$\downarrow$} & {tLPIPS$\downarrow$} \\
    \midrule
    Without two-stage Heun solver & 0.586 & 0.135 & 0.238 \\
    With two-stage Heun solver  & \textbf{0.603} & \textbf{0.121} & \textbf{0.224} \\
    \bottomrule
  \end{tabular}
\caption{Effect of two-stage Heun solver}
\label{tab:heun-solver}
\end{table}

\noindent
\textbf{Effect of Window Length}. We experimented with shorter window lengths (e.g. 50 frames) processed by the model by artificially limiting the temporal attention. Shorter windows mean more frequent blending, which could accumulate error but also could refresh context. We found 80–100 frame windows to be optimal for our model; very short windows (30 frames) hurt global consistency (some long-term context was lost). Using the model's suggestion (81) was a safe choice.

These ablations confirm that each design choice contributes to the robust performance of our system.

\section{Limitations}
Our approach inherits limitations of the underlying T2V backbone: extremely long-range identity persistence and fine-grained geometry may still drift under large edits or very high-motion scenes, and the two-stage solver roughly doubles per-step compute compared to first-order sampling. In practice, the improved stability allows fewer total diffusion steps, partially offsetting this cost.
In addition, because edited regions lack ground-truth, our quantitative protocol focuses on background preservation, prompt alignment, and temporal stability; richer human preference studies are left to future work.

\section{Conclusion}
We presented a unified framework to achieve long-form video inpainting and outpainting by combining LoRA-based fine-tuning with an overlapping high-order diffusion sampling strategy. Starting from the Wan 2.1 foundation model, we turned it into a flexible, high-quality video editor capable of filling in or extending content over hundreds of frames. Through the dual-region loss and mask conditioning, the LoRA adaptation preserves the original content while seamlessly painting missing regions – all without modifying the original model architecture. Through overlapping window denoising and second-order solver integration, we scale the generation to substantially longer durations with smoother transitions and reduced boundary artifacts between segments.
Our experiments show improved fidelity, perceptual quality, and temporal stability over strong baselines on long-form inpainting and outpainting. Beyond these tasks, the same LoRA adaptation and overlapping high-order sampling strategy can be applied to other mask-guided or control-guided video edits with minimal additional training.



\bibliography{IJCNN2026}
\bibliographystyle{IEEEtran}

\end{document}